\documentclass[10pt,twocolumn,letterpaper]{article}

\usepackage[pagenumbers]{cvpr} 

\usepackage[dvipsnames]{xcolor}

\definecolor{cvprblue}{rgb}{0.21,0.49,0.74}
\usepackage[pagebackref,breaklinks,colorlinks,citecolor=cvprblue]{hyperref}
\usepackage{times}
\usepackage{epsfig}
\usepackage{graphicx}
\usepackage{amsmath}
\usepackage{amssymb}
\usepackage{booktabs}
\usepackage{multirow}
\usepackage{enumerate}
\usepackage{float}
\def\confName{CVPR}
\def\confYear{2024}

\title{RobustCalib: Robust Lidar-Camera Extrinsic Calibration with \\ Consistency Learning}

\author{Shuang Xu$^{1}$, Sifan Zhou$^{1,2}$\thanks{Work done as an intern at Meituan.}, Zhi Tian$^{1}$, Jizhou Ma$^{1}$, Qiong Nie$^{1}$, Xiangxiang Chu$^{1}$\thanks{Corresponding author.}
\\
{\small $^{1}$ Meituan Inc.}
{\small $^{2}$ Southeast University}
\\
 {\tt\small
\{xs19971114,sifanjay\}@gmail.com, zhi.tian@outlook.com,} \\
{\tt\small\{majizhou,nieqiong,chuxiangxiang\}@meituan.com}}
\begin{document}
\maketitle
\begin{abstract}
   In autonomous vehicles and robotic systems, 3D LiDARs and 2D cameras are widely used for environment perception. The fusion of LiDAR and camera has shown impressive performance gains in various perception tasks. One important basis of LiDAR-camera fusion is the accurate extrinsics between them. 

Current traditional methods for LiDAR-camera extrinsics estimation depend on offline targets and human efforts, while learning-based approaches resort to iterative refinement for calibration results, posing constraints on their generalization and application in on-board systems. In this paper, we propose a novel approach to address the extrinsic calibration problem in a robust, automatic, and single-shot manner. Instead of directly optimizing extrinsics, we leverage the consistency learning between LiDAR and camera to implement implicit re-calibartion. Specially, we introduce an appearance-consistency loss and a geometric-consistency loss to minimizing the inconsitency between the attrbutes (\textit{e.g.,} intensity and depth) of projected LiDAR points and the predicted ones. This design not only enhances adaptability to various scenarios but also enables a simple and efficient formulation during inference. We conduct comprehensive experiments on different datasets, and the results demonstrate that our method achieves accurate and robust performance. To promote further research and development in this area, we will release our model and code.


\end{abstract}   

\section{Introduction}
\label{sec:intro}
Robust and accurate perception is a crucial issue in self-driving and robotics. Due to the complexity and high dynamic of the self-driving environment, using a single sensor can not guarantee robust perception ability in various scenes. Conversely, multi-sensor fusion (such as LiDAR and camera) can effectively improve the accuracy and robustness of perception in real-world environments with the complementary nature of the two sensors \cite{bai2022transfusion, liu2023bevfusion, yoo20203d, liang2018deep}.In detail, LiDAR provides 3D geometrics and reflection intensity information within sparse point cloud representation, which allows for accurate distance measurements. Besides, LiDAR is more robust against illumination change. Differently, cameras capture abundant color and texture features within dense RGB images, which provides more fine-grained details.

\begin{figure}[t]
  \centering
  \includegraphics[width=\linewidth]{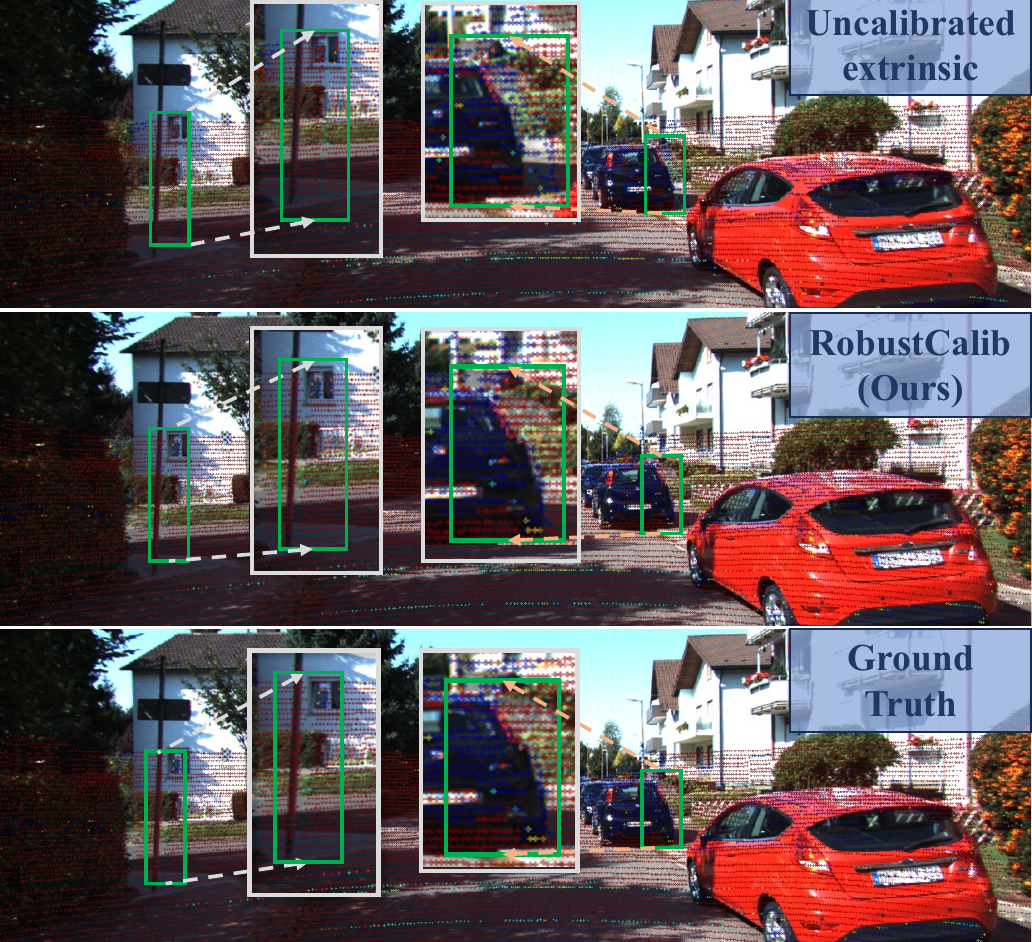}
  \caption{Projection of lidar points onto image colored with intensity information. \textbf{(Top:)} Uncalibrated extrinsic. \textbf{(Middle:)} Corrected extrinsic using our proposed method. \textbf{(Bottom:)} Ground truth. The image with ID 003837 is in the 00 sequence of the KITTI dataset. Best viewed in color.}
  \label{fig:frame_work}
\end{figure}

The combination of LiDAR and camera provides the feasibility to overcome the inherent limitations of each sensor. One important basis of fusing these two heterogeneous sensors lies in the precise extrinsic parameter estimation (relative rigid body transformation between them). 

Early target-based calibration methods \cite{geiger2012automatic, park2014calibration, pusztai2017accurate} primarily rely on checkboard or specific calibration plates towards accurate results. However, these methods endure notorious drawbacks, such as time-consuming, labor-intensive, environment-depended, etc. As a result, these methods are not suitable for large-scale industrial applications. Unlike target-based methods, some target-less methods \cite{yuan2021pixel, koide2023general} have been explored for online automatic calibration. However, these methods often have high requirements for geometric features in the environment and may lack generalizability.

Considering the remarkable success archived by neural networks in computer vision tasks, some deep learning-based methods are introduced for multi-sensor extrinsics estimation. RegNet \cite{schneider2017regnet} is the first introduced learning-based method, which employs a multi-scale iterative refinement formulation. Following RegNet, various methods \cite{lv2021lccnet, iyer2018calibnet} are built upon this formulation during training and inference. This complex formulation inevitably poses challenges for deployment on autonomous vehicles as well as robotic systems. In addition, current methods mostly evaluate their performance on KITTI scenes using only one camera, leaving room for improvement in the performance and generalization ability.

In this paper, we introduce a novel paradigm to address the challenge of extrinsic calibration between LiDAR and camera systems. Different from previous methods that depend on specific scene features or improve calibration performance through iterative refinement, our proposed method does not impose assumptions on environmental priors, and generates the corrected extrinsics in an end-to-end, single shot manner. For the first time, we transform the direct extrinsics optimization problem into a consistency learning problem between LiDAR and cameras. By minimizing the inconsistency between the projected point cloud attributes (\emph{e.g.}, appearance attributes as intensity, geometric attributes as depth) and the predicted ones, our PoseNet is implicitly enforced to predict accurate extrinsic parameters. Furthermore, this novel design enables a concise and deployment-friendly subnetwork during inference. Our method demonstrates that with just the input of paired LiDAR and camera data, along with the initial, uncalibrated extrinsics, our proposed method automatically outputs the recalibrated extrinsics.

To summarize, our main contributions are three-fold:
\begin{itemize}
\item {We propose a consistency constraint to enforce the appearance-consistency and geometric-consistency of LiDAR and camera, leading to an automatic extrinsic calibrator across multiple sensors.}
\item {We unveil a novel paradigm for sloving the SE(3) pose problem. The proposed RobustCalib is concise in its formulation, requiring only a subnetwork during inference phase and eliminating the need for iterative refinement. This characteristic is particularly beneficial for industrial applications, where efficiency and reliability are paramount.}
\item {Extensive experiments on the KITTI, nuScenes and MT-ADV datasets demonstrate the superior performance and robust generalization ability of our RobustCalib. These results affirm its applicability and effectiveness in diverse real-world scenarios.}
\end{itemize}

\section{Related Work}
\label{sec:formatting}

\subsection{Target-based Methods}
Traditional target-based calibration methods \cite{geiger2012automatic, park2014calibration, pusztai2017accurate, wang2017reflectance, kim2020extrinsic, an2020geometric} need certain environment and equipment, such as checkerboards or custom-made markers placed on the walls. Mishra et al. \cite{EC3D2020} proposed a LiDAR-camera extrinsic estimation method on a marker-less planar target by utilizing the geometric constraints between planes and lines. Toth et al. \cite{toth2020automatic} presented an automatic calibration method using spheres whose contours can be accurately detected on LiDAR and camera. Recently, Yan et al. \cite{yan2023joint} located the LiDAR pose by adding circular holes around the checkerboard, then using the reprojection constraint to joint calibrate camera intrinsic and LiDAR-camera extrinsic. Target-based methods are accurate and commonly used. However, these methods require tuning a lot of parameters and iterative optimization, which are time-consuming and costly to implement regularly on a vehicle.


\subsection{Target-less Methods}
Target-less methods \cite{tamas2013targetless, tamas2014relative, pandey2015automatic, taylor2015multi, kang2020automatic} usually use general geometric features (such as lines or edges) from natural scenes to calibrate the extrinsic parameters based on geometric constraints. Yuan et al. \cite{yuan2021pixel} adopted point cloud voxel cutting and plane fitting for LiDAR edge extraction to achieve automatic extrinsic calibration in targetless environments. Recently, Koide et al. \cite{koide2023general} presented a fully automatic LiDAR-camera calibration toolbox without targets. They first estimate the LiDAR camera transformation via RANSAC \cite{fischler1981ransac}, and then refine the transformation based on the normalized information distance. These target-less methods are automated and can be computed online. However, they still rely on computation-heavy optimization and require feature-rich environments.

\begin{figure*}[!htb]
  \centering
  \includegraphics[width=\textwidth]{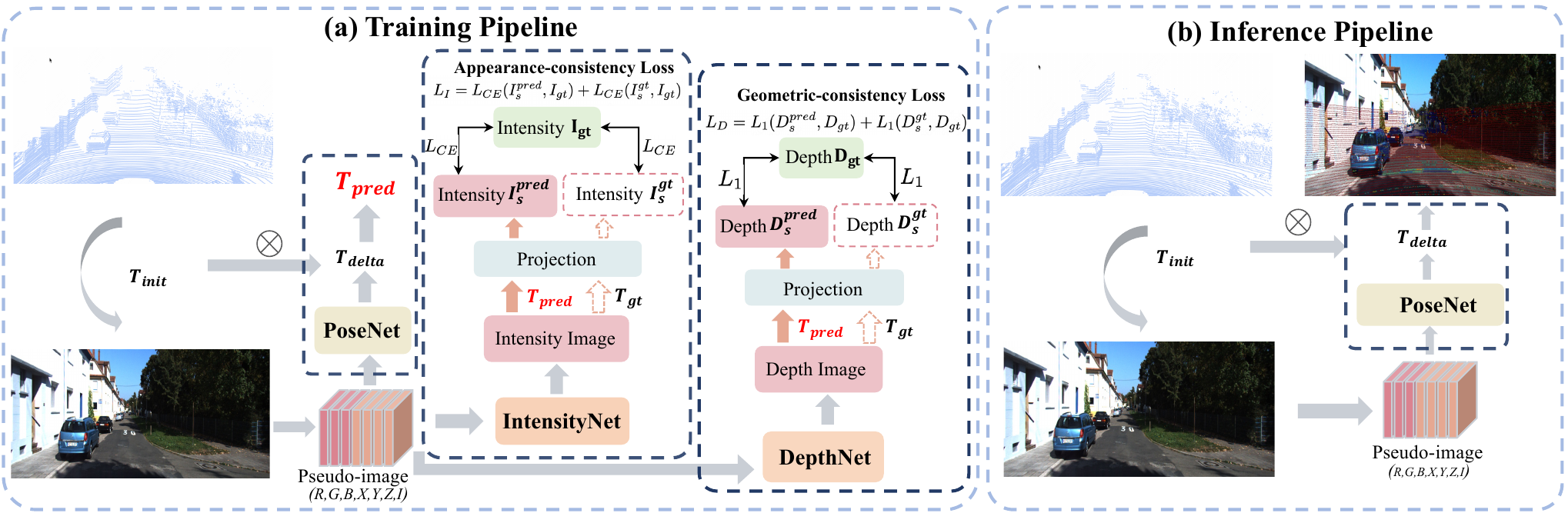}
  \caption{The detailed architecture of the proposed RobustCalib. \textbf{(a)} The training pipeline. \textbf{(b)} The inference pipeline. Once the training process is completed, we only use \textbf{PoseNet} for inference to obtain the calibration extrinsic parameters, which enables point cloud and image alignment. Best viewed in color.}
  \label{fig:frame_work}
\end{figure*}

\subsection{Learning-based Methods}

Considering the recent success of deep learning, learning-based methods \cite{schneider2017regnet, iyer2018calibnet, 2019cmrnet, yuan2020rggnet, wang2020soic, zhu2020online, 2022I2D-Loc} often take advantage of CNN to estimate the extrinsic parameters between LiDAR and cameras. RegNet \cite{schneider2017regnet} is a pioneering work, which leverages CNN to predict the 6-DoF extrinsic parameters. CalibNet \cite{iyer2018calibnet} is another significant work, which maximizes the geometric and photometric consistency between the input image and the point cloud to estimate the 6-DoF extrinsic parameters in real-time. After that, RGGNet \cite{yuan2020rggnet} proposes a deep generative model to construct a tolerance-aware loss considering the Riemannian geometry. NetCalib \cite{2020NetCalib} optimizes the geometric error between the depth map from stereo cameras and the depth map from LiDAR projected to achieve more accurate performance. Besides, Zhu et al. \cite{zhu2020online} formulated extrinsic calibration as an optimization problem. They proposed a calibration quality metric based on semantic features to align pairs of camera and LiDAR frames. LCCNet \cite{lv2021lccnet} is a milestone work, which adopts the cost-volume \cite{sun2018pwc} as its backbone and achieves good performance via multi-scale refinement. Recently, Alexander et al. \cite{2022ECCSS} achieved extrinsic calibration by registering image and point cloud with semantic labels after semantic segmentation. These methods achieve accurate calibration but leave room for improvement in accuracy, performance, and reliability in unstructured environments. In this paper, we make the first attempt to explore the consistency-learning to achieve more accurate and robust online LiDAR-camera extrinsic calibration. Different from previous works, the porposed method can be generalized to various practical application scenarios with a simple yet efficient inference structure.

\section{Method}

\subsection{Method Overview}
Our goal is to train a learning-based network for online lidar-camera extrinsic calibration. Leveraging constrainting the consistency (\textit{i.e.}, appearance-consistency and geometric-consistency) between lidar and camera, the proposed method is totally end-to-end, avoiding multiple refinements during inference. The pipeline of RobustCalib is illustrated in Fig. \ref{fig:frame_work}, RobustCalib consists of three sub-networks: \textbf{PoseNet}, which predicts the extrinsic $T_{pred}$ between the LiDAR and camera; \textbf{IntensityNet}, which predicts pseudo intensity image; \textbf{DepthNet}, which predicts depth. Given an RGB image $I$, a point cloud $P$, and initial extrinsic $T_{init}$ as input. Firstly, we project the point cloud into the pixel coordinate system to get a unified pseudo-image representation with 7 channels (\textit{i.e.}, RGB channels $r,g,b$, camera coordinates $ x_i, y_i, z_i $ and intensity $I_i$). Secondly, we input this pseudo-image to PoseNet, IntensityNet, and DepthNet. In PoseNet, we input the pseudo-image and get the delta extrinsic prediction. After that, we multiply the extrinsic prediction with the initial $T_{init}$ to obtain the calibrated extrinsic $T_{pred}$. In IntensityNet, after getting the predicted pseudo-intensity image, we respectively project the predicted intensity using the $T_{pred}$ and $T_{gt}$ to construct constraints based on correspondence and learning appearance-consistency between liDAR and camera. Similarly, in DepthNet, after getting the predicted pseudo-depth image, we respectively project the predicted depth using the $T_{pred}$ and $T_{gt}$ to construct constraints based on geometric-consistency between liDAR and camera. In the following sections, we discuss the data preparation and representation, the core idea of lidar-camera consistency learning, and the details of network architecture.


\subsection{Data Preparation and Representation}
The performance of data-driven learning-based methods improve with the quality and quantity of data. In this case, we follow previous methods (\textit{i.e.}, RegNet \cite{schneider2017regnet}, LCCNet \cite{lv2021lccnet} and UniCal \cite{cocheteux2023unical}) which add random de-calibration within a reasonable range to the ground truth extrinsic calibration parameters $T_{LC}$. Thus generating the initial extrinsic $T_{init} = \Delta T \cdot T_{LC} $ and mitigating insufficiency of training data.
In practical autonomous driving operation scenarios, ranging $\pm 10cm$ for translation and $\pm 1^\circ$ for rotation is an empirical estimate of the levels of perturbation, which is also reported in UniCal \cite{cocheteux2023unical}. Thus we choose the same translation and rotation range $\left\{  \pm 10cm, \pm 1^\circ \right\}$.

Considering the inherent different characteristics of multimodal data (image and point cloud in our case), the efficient representation and fusion of them is an inevitable challenge. In this paper, we formulate the image input and the point cloud input into a unified pseudo-image representation. 
Specifically, let $I \in \mathbb{R}^{H \times W \times 3}$ be an RGB image and $P \in \mathbb{R}^{N \times 4 }$ be a point cloud. And $P_{i} = \left[ X_i, Y_i, Z_i, I_i \right] $, the first three dimensions represent spatial coordinates of a point, the last dimension $I_i$ represents intensity. To establish correspondences between multimodal inputs, we first project lidar point onto camera's coordinate system  $c_{i} = \left[ x_i, y_i, z_i \right] $ and pixel coordinate system $p_{i} = \left[ u_i, v_i \right] $, using the initial calibration ${T}_{init}$ and the camera intrinsic $K$. The projection process can be formulated as follows:
\begin{equation}
    {{c}_{i}^{T}}= {{T}_{init}} \cdot {{P}_{i}^{T}}
\end{equation}

\begin{equation}
    {{z}_{i} \cdot {p}_{i}^{T}}= {K \cdot {T}_{init}} \cdot {{P}_{i}^{T}}
\end{equation}

To obtain the projection of point cloud, we generate an $N$ channel pseudo-image as network input. In our implementation, $N$ is set to 7, which contains 3 channels for RGB information, and 4 channels for lidar information (\textit{i.e.}, intensity $I_i$ and camera coordinates $ x_i, y_i, z_i $). 

\subsection{Lidar-Camera Consistency Learning}
As mentioned above, we design the consistency constraint (\textit{i.e.}, appearance-consistency and geometric-consistency) to ensure alignment between the lidar and camera, which implicitly optimizes the extrinsic between them. Here, we present the core idea and advantages of our design.

\begin{figure*}[!htb]
  \centering
  \includegraphics[width=\textwidth]{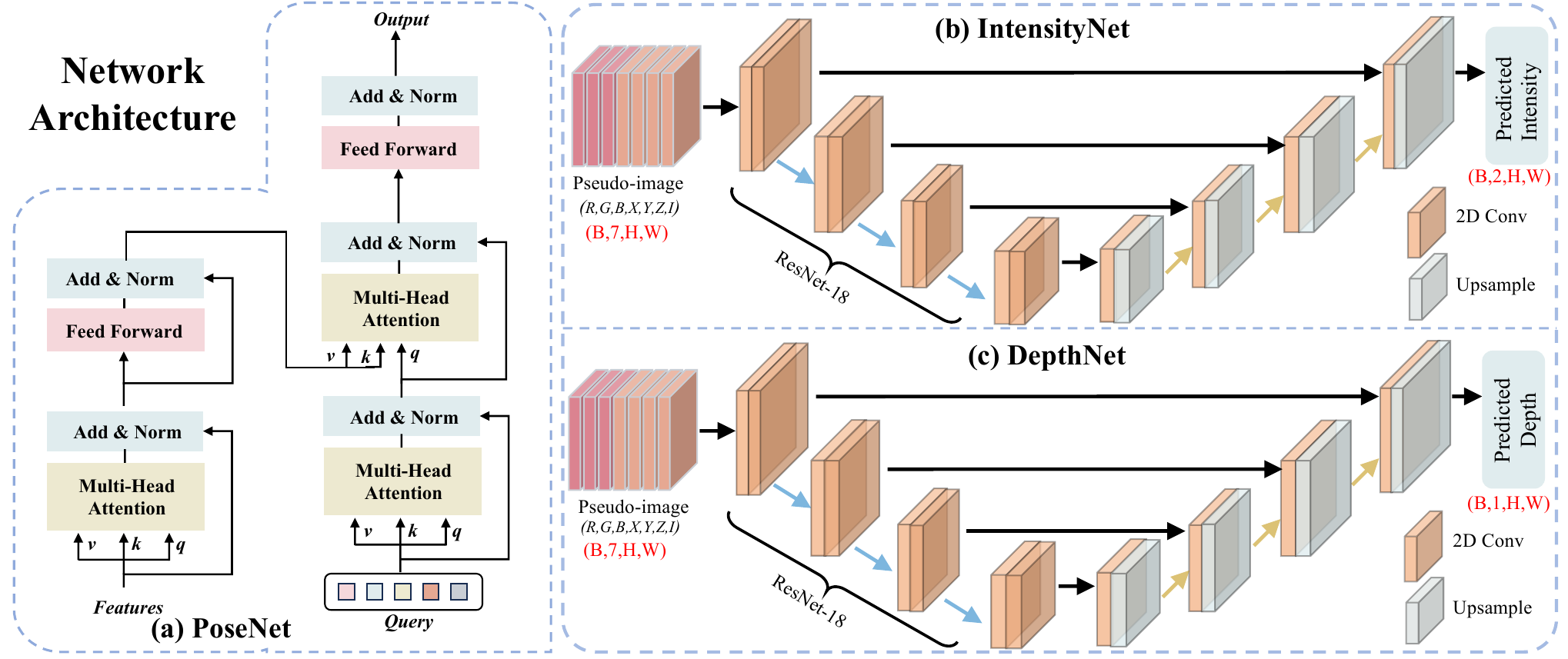}
  \caption{(a) Architecture of the transformer decoder layers of the PoseNet. (b) Architecture of the IntensityNet. (c) Architecture of the DepthNet. }
  \label{fig:network}
\end{figure*}

\textbf{Appearance-consistency Loss.} Takes the unified representation as input, we first feed it into an encoder-decoder like IntensityNet. The detailed architecture of IntensityNet will be introduced in the subsection \ref{Architecture}. After feature encoding, the IntensityNet aims to predict a pseudo intensity image. In order to simplify optimization and accelerate convergence, we transform the intensity regression task into a binary classification task. That is, the intensity is mapped to 1 or 0 depending on whether the value is greater than a threshold. The setting of the intensity threshold will be specifically elaborated in Section \ref{sec:Implementation Details}.
Correspondingly, the predicted pseudo intensity image is shaped as $C \times H \times W $ (C is set as 2), which is obtained by an intensity binary classifier. 

The PoseNet parallelly takes the unified representation as input, and outputs extrinsic in the format of translation in meters and rotation in Euler angles. Then the output is transformed into a rotation translation matrix $T_{delta}$, multiplied with $T_{init}$, and obtained the $T_{pred}$. The detailed architecture of PoseNet is illustrated in subsection \ref{Architecture}. In order to construct correspondence and learning appearance-consistency between lidar and camera, we respectively project the point cloud into the pixel coordinate system using the $T_{pred}$ and $T_{gt}$, and sample values from the predicted pseudo intensity image. The sampled intensity denoted as $I^{pred}_{s}$ and $I^{gt}_{s}$ respectively and supervised by the binary intensity label $I_{gt}$ via cross-entropy loss. This process can be formulated as: 
\begin{equation}
    {L_{I}}= L_{CE}(I^{pred}_{s}, I_{gt}) + L_{CE}(I^{gt}_{s}, I_{gt})
\end{equation}

\textbf{Geometric-consistency Loss.} Formulated as a similar paradigm of the appearance-consistency learning, the geometric-consistency learning takes depth as the medium. The DepthNet, which is also an encoder-decoder structure, predicts a one-channel pseudo depth image. Then the LiDAR points are projected using $T_{pred}$ and $T_{gt}$, and the corresponding depth values are sampled by the pixel coordinates, denoted as $I^{pred}_{s}$ and $I^{gt}_{s}$, respectively. We take the depth label (denoted as $D_{gt}$) obtained by the extrinsic ground truth projection as the learning target and formulate the objective function as:
\begin{equation}
    {L_{D}}= L_{1}(D^{pred}_{s}, D_{gt}) + L_{1}(D^{gt}_{s}, D_{gt})
\end{equation}

\textbf{User-friendly Inference Model.} Notably, in the training process, we use the appearance-consistency and geometric-consistency to constrain the extrinsic prediction. However, in the inference stage, IntensityNet and DepthNet are not needed. The inference pipeline is concise and formulated in a single-branch pipeline.  Our inference process only takes the lidar-camera information and initial extrinsic $T_{init}$ as input, and automatically outputs the corrected extrinsic $T_{pred}$ in a single-shot manner, which shows advantages in application compared with previous iterative refinement methods \cite{schneider2017regnet, lv2021lccnet}.


\subsection{Network  Architecture} 
\label{Architecture}

\indent \textbf{PoseNet.} The PoseNet parallelly takes the unified representation as input, and outputs extrinsic in the format of translation in meters and rotation in Euler angles. For PoseNet, we formulate it as a ResNet-50 \cite{he2016resnet} based encoder and transformer-based \cite{transformer} decoder architecture. The decoder part takes the encoded feature ($B \times C \times h \times w $, and reshaped as $B \times N_{1}\times C $) from lidar and image, and the randomly initialized queries (shaped as $B \times N_{2}\times C$) as inputs. We first compute self-attention score of features and queries separately, as well as cross-attention score between them. Then the queries equipped with calibration information are decoded into query features (shaped as $B \times N_{1}\times C $) by a feed-forward network (FFN). Two parallel Linear layers are adopted to transform the query feature to three-channel outputs that respectively represent rotation in Euler angles and translation in meters. Finally, the rotation and translation are represented as a rotation translation matrix $T_{delta}$, which is multiplied with the initial $T_{init}$ to obtain the corrected extrinsic $T_{pred}$. The architecture of PoseNet, IntensityNet, and DepthNet can be seen in Fig. \ref{fig:network}.

\textbf{IntensityNet.} The IntensityNet is an encoder-decoder architecture. The encoder part is based on ResNet-18 \cite{he2016resnet}, and the first convolutional layer is adjusted to adapt the unified input with $N$ channels. The decoder part is composed of convolutional-based up-sampling layers, gradually restoring the encoded features to the original resolution. The final classifier is configured to output two-channel intensity prediction ($B \times 2 \times h \times w $). 

\textbf{DepthNet.} The DepthNet enjoys a similar architecture as the IntensityNet, the only difference is that the final classifier is adjusted to predict one-channel depth ($B \times 1 \times h \times w $).

\begin{figure*}[!htb]
  \centering
  \includegraphics[width=0.9\textwidth]{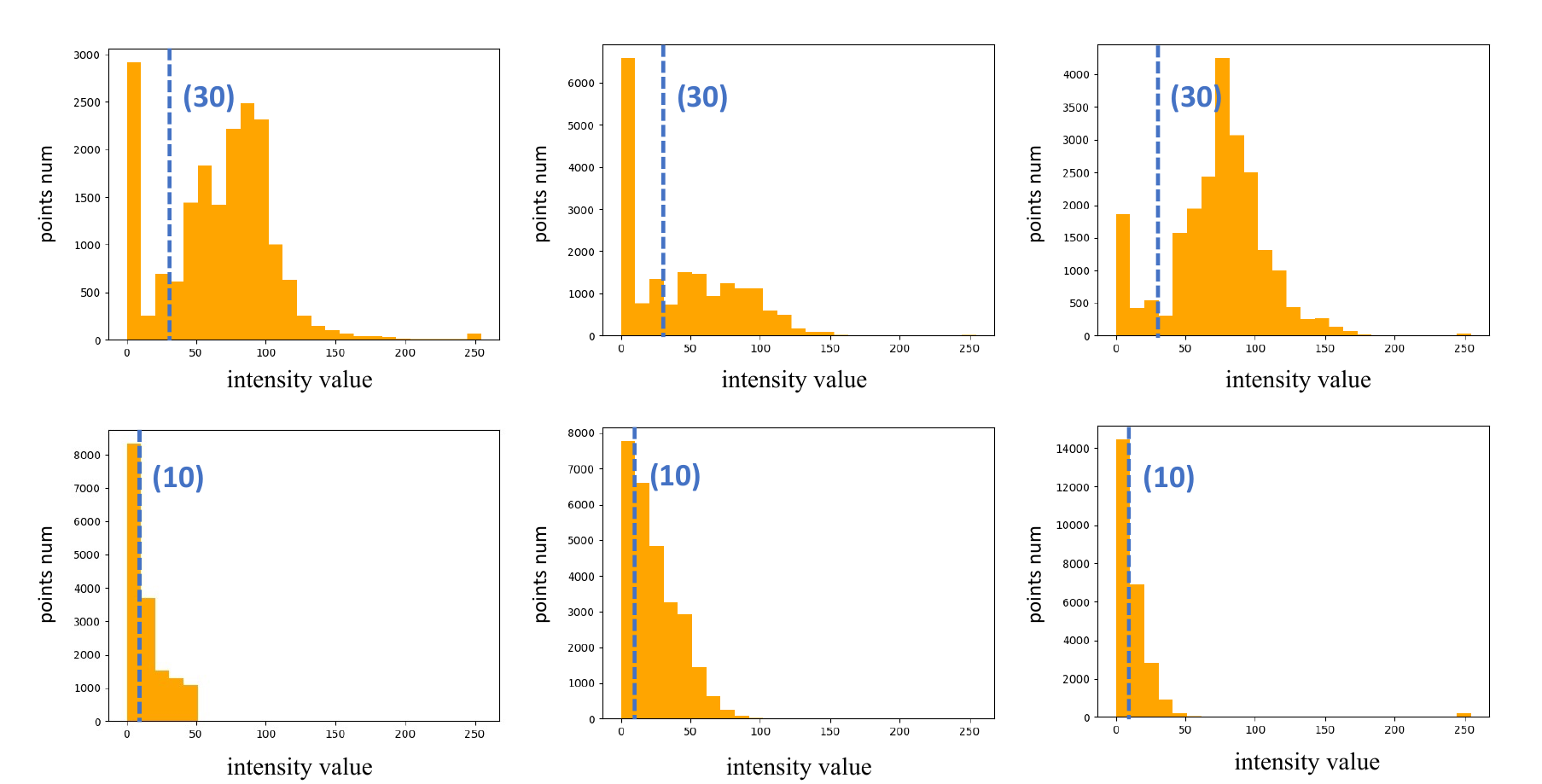}
  \caption{Illustrations of intensity distribution in both the KITTI and the MT-ADV dataset. The first row is data collected from the KITTI dataset, with frames from sequence 18, 21, and 02 arranged from left to right. The second row is data sampled from the MT-ADV dataset.}
  \label{fig:thres_intensity}
\end{figure*}

\section{Experiments}
\label{sec:experiments}
In this section, we evaluate our proposed calibration approach on the KITTI odometry dataset, nuScenes dataset and MT-ADV dataset. We detail the data processing, training details, quantitative experiments, ablation study and visualization results.

\subsection{Dataset}
\paragraph{KITTI dataset.} KITTI \cite{geiger2012kitti} is widely used as a reference dataset in current mainstream calibration methods. In this paper, we use the odometry branch of the KITTI dataset to verify the proposed method. KITTI Odometry dataset consists of 21 sequences from different scenarios using 64-line Velodyne 3D LiDAR (10 HZ) and camera with a resolutions of 1392 $\times$ 1032, and provides accurate ground truth extrinsic calibration parameters. Following LCCNet \cite{lv2021lccnet} setting, we split the KITTI dataset as follows: 01-21 for training (39011 frames) and 00 for test (4541 frames).

\paragraph{nuScenes dataset.} We also conducted experiments on the nuScenes dataset ~\cite{caesar2020nuscenes}, which is a large-scale autonomous driving dataset. Specifically, the perception system in nuScenes consists of a 32-line LiDAR and 6 surround-view cameras with a resolutions of 1600 $\times$ 900 and provides accurate ground truth extrinsic calibration parameters. nuScenes dataset contains 1640310 frames for training and 351006 frames for test from different scenarios.


\paragraph{MT-ADV dataset.} To further evaluate the generalization ability and robustness of the proposed method, we also conducted experiments on the Meituan autonomous delivery vehicles (MT-ADV) dataset. Specifically, the perception system in MT-ADV consists of a 64-line LiDAR and 6 surround-view cameras. MT-ADV dataset contains 543057 frames for training and 87904 frames for test. The ground truth extrinsics between LiDAR and cameras are generated from Meituan's calibration infrastructure. Unlike the KITTI dataset, MT-ADV dataset includes more complex real-word scenarios and a larger number of data samples, which also brings more difficulties and challenges. Besides, MT-ADV dataset encompasses multiple cameras, which more meets the requirements of the practical application scenarios.

\subsection{Implementation Details.}
\label{sec:Implementation Details}
\paragraph{Training details.} The proposed method is trained totally end-to-end. We use SGD Optimizer during training with the parameter weight decay set as $0.0001$ and momentum set as $0.9$. We train the network with an initial learning rate $0.01$ and adopt the consine decay strategy. For the KITTI dataset, the network is trained for $240,000$ iterations. 

\paragraph{Intensity threshold setting.} In the section of apperance-consistency learning, we formulate the intensity regression problem as a binary classification problem. Therefore, there needs a reasonable threshold to classify the LiDAR points depending on their intensity value. In our experiment, we set this threshold based on the statistic of the intensity distribution in different datasets. Here we present several examples intensity distribution in the KITTI dataset and the MT-ADV dataset in Fig. \ref{fig:thres_intensity}. The first row is collected from the KITTI dataset, the second row is collected from the MT-ADV dataset. Based on the intensity distribution of LiDAR points, we can see that 30 is a reasonable threshold for split in KITTI, as well as 10 for the MT-ADV dataset.

\begin{table*}[t]
\begin{center}
\begin{tabular}{c|cccc|cccc}
\toprule
\multirow{2}{*}{Method}   & \multicolumn{4}{c}{Translation absolute Error (cm)} & \multicolumn{4}{c}{Rotation absolute Error (${}^\circ $)} \\
   & mean  & X  & Y  & Z   & mean  & Roll  & Pitch & Yaw \\
\midrule
Regnet ~\cite{schneider2017regnet}  & 6.00 & 7.00 & 7.00  & 4.00   & 0.28 & 0.24 & 0.25 & 0.36  \\ 
Calibnet \cite{iyer2018calibnet} & 4.34 & 4.20 & 1.60 & 7.22 & 0.41 & 0.18 & 0.90 & 0.15 \\ 
NetCalib \cite{2020NetCalib} & 1.69 & 1.20 & 2.77 & 1.10  &0.09 & 0.04 & 0.16 & 0.09  \\ 
LCCNet~\cite{lv2021lccnet} & 1.292 & - & - & -  & 0.173 & - & - & - \\
DXQ-Net~\cite{jing2022dxqnet} &1.425 &0.754 &0.476 &1.091  &0.084 &0.049 &0.046 &0.032  \\
\hline
RobustCalib (\textbf{ours}) &\textbf{0.806} &0.820  &0.740  &0.860   &\textbf{0.046}&0.034&0.037&0.068 \\ 
\bottomrule
\end{tabular}
\end{center}
\caption{Comparison with current state-of-the-art deep learning-based calibration methods on KITTI dataset.}
\label{tab:compare-kitti}
\end{table*}

\begin{figure*}[!htb]
  \centering
  \includegraphics[width=\textwidth]{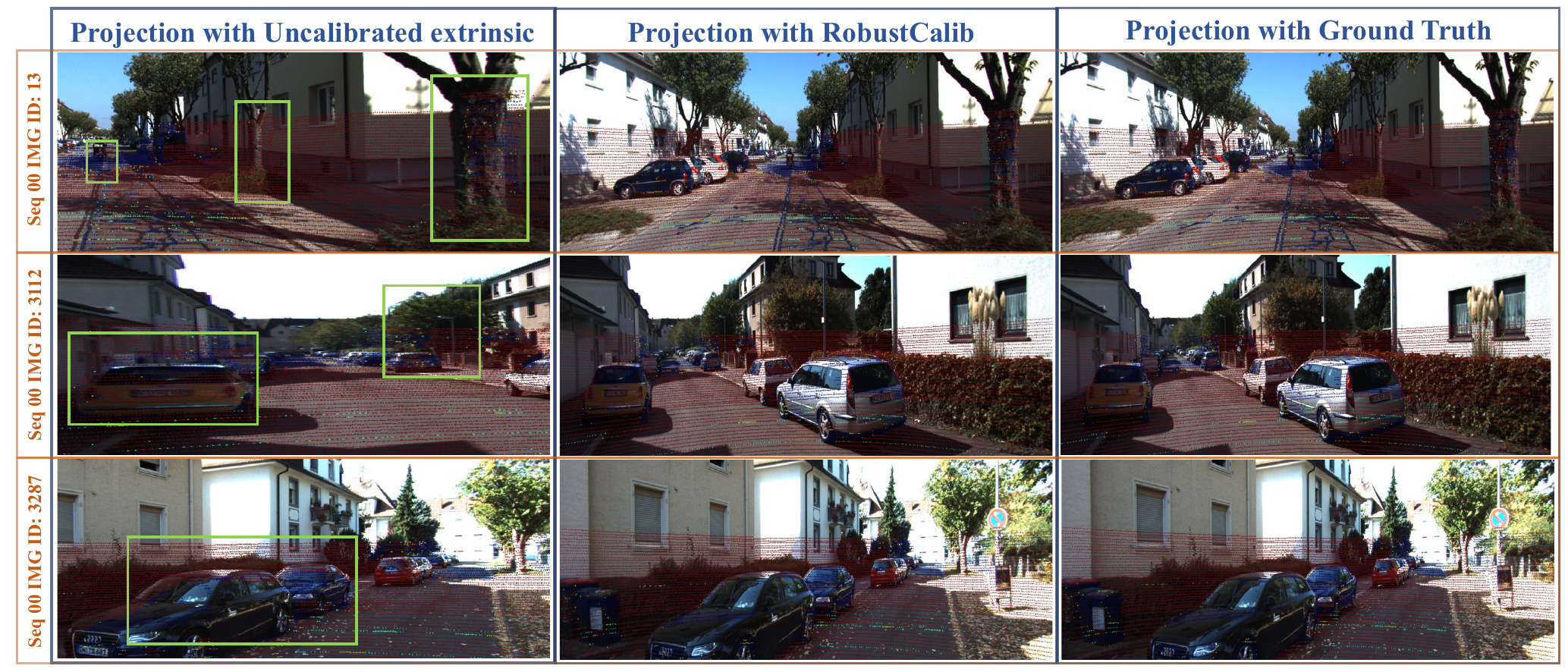}
  \caption{Visualization of calibration results from different scenes of KITTI dataset. The first column shows the projecion of LiDAR points onto the image using the uncalibrated extrinsics. The second column shows the calibrated extrinsics produced by our method. The third column shows the ground truth results. The \textcolor{green}{green box} show the mis-alignmnet. Better view zoom in with color.}
  \label{fig:kitti_res}
\end{figure*}

\begin{table*}[!htb]
\begin{center}
\begin{tabular}{c|cccc|cccc}
\toprule
\multirow{2}{*}{Method}   & \multicolumn{4}{c}{Translation absolute Error (cm)} & \multicolumn{4}{c}{Rotation absolute Error (${}^\circ $)} \\
   & mean  & X  & Y  & Z   & mean  & Roll  & Pitch & Yaw \\
\midrule
LCCNet~\cite{lv2021lccnet} & 2.218 & 2.430 & 2.176 & 2.048  & 0.154 & 0.149 & 0.124 & 0.189 \\
\hline
RobustCalib (\textbf{ours}) &\textbf{1.600} &2.000  &1.530  &1.270   &\textbf{0.114}&0.103&0.085&0.155 \\ 
\bottomrule
\end{tabular}
\end{center}
\caption{Comparison with state-of-the-art learning-based calibration methods on nuScenes dataset.}
\label{tab:compare-nus}
\end{table*}

\begin{figure*}[!htb]
  \centering
  \includegraphics[width=0.9\textwidth]{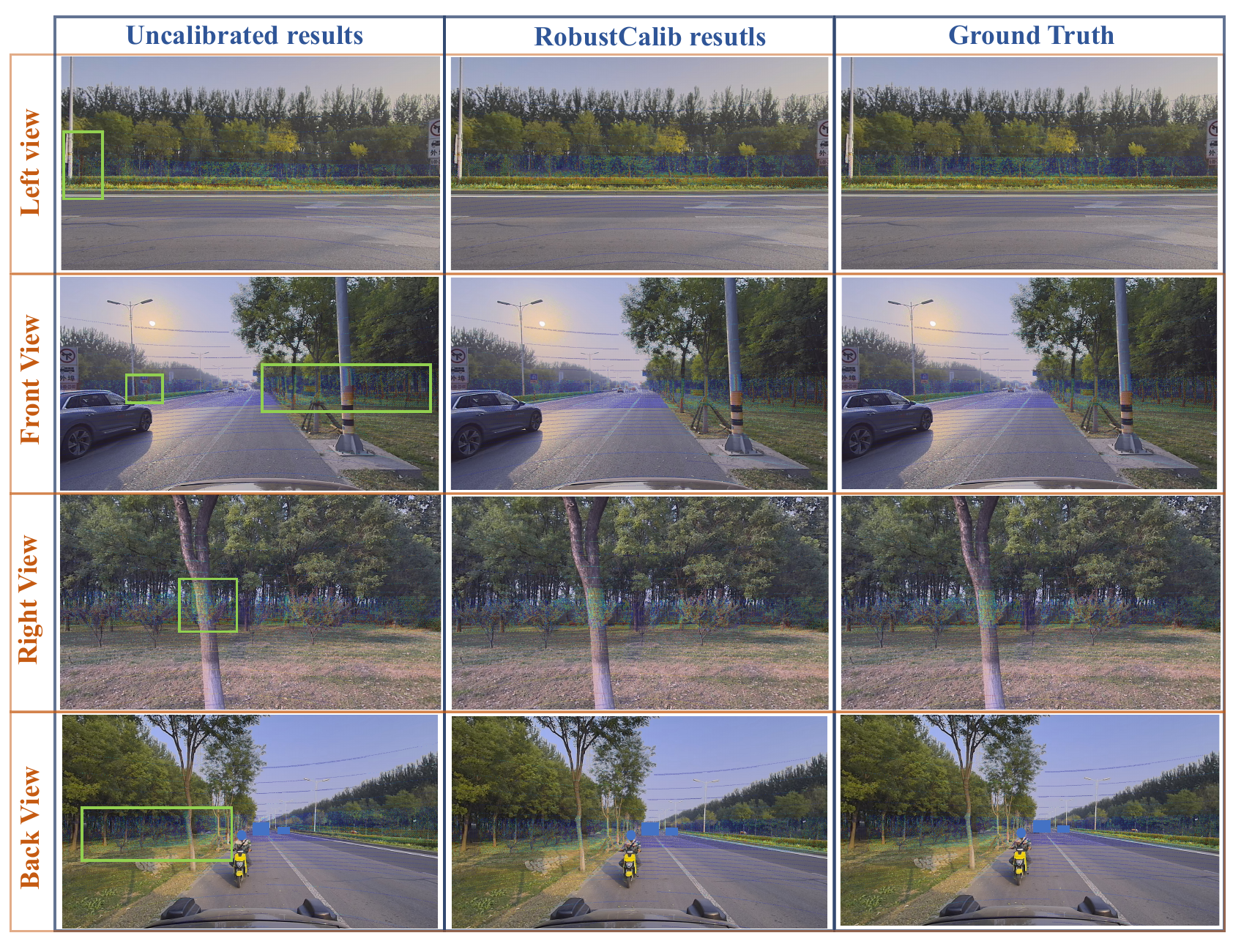}
  \caption{Visualization of calibration results under multiple views from MT-ADV dataset. The first column shows the projecion of LiDAR points onto the image using the uncalibrated extrinsics. The second column shows the calibrated extrinsics produced by our method. The third column shows the ground truth results. The \textcolor{green}{green box} show the mis-alignmnet. Better view zoom in with color.}
  \label{fig:MT_res}
\end{figure*}

\subsection{Evaluation Metrics}
In this section, we introduce the evaluation metrics for calibration parameters. The metric of mean absolute error is used for translation and rotation.  For translation, we compute the translation verctor's absolute error in X, Y, Z directions and the mean value. For rotation, we transform the rotation metric to the formulation of Euler angles and compute the angle error of Roll, Pitch, Yaw axes.

\subsection{Overall Results}
\subsubsection{Results on KITTI.} 
\label{sec:kitti-ex}
The calibration results for our proposed RobustCalib on KITTI are shown in Table \ref{tab:compare-kitti}. We can observe that RobustCalib achieves a mean translation error of 0.806cm (x, y, z: 0.820cm, 0.740cm, 0.860cm), a mean angle error of $0.046^\circ$ (roll, pitch, yaw: $0.034^\circ$, $0.037^\circ$, $0.068^\circ$), which is a new state-of-the-art performance in both translation and rotation metrics. Compared to the state-of-the-art method DXQNet~\cite{jing2022dxqnet}, our method achieves a nearly 50\% reduction in both mean translation error and mean angle error. Compared to LCCNet~\cite{lv2021lccnet}, the mean angle error of RobustCalib is reduced by approximately 3 times. It is worth noting that our RobustCalib favorably outperforms other learning-based methods with a single shot manner, while other methods (\textit{i.e.}, RegNet \cite{schneider2017regnet}, LCCNet \cite{lv2021lccnet} and UniCal \cite{cocheteux2023unical}) rely on iterative refinement during training and inference process.

As shown in Fig.\ref{fig:kitti_res}, we visualize the calibration results of different scenarios on KITTI dataset. Here, the LiDAR points are colored with intensity values. It can be seen that our proposed RobustCalib is capable of providing re-calibrated extrinsic parameters that align the LiDAR point cloud projection onto the corresponding camera images. Particularly, for the obvious uncalibrated results (\textcolor{green}{green box} in 1st column), after applying our RobustCalib, achieving the almost same results as the ground truth (2nd and 3rd columns in Fig.\ref{fig:kitti_res}).

\begin{table*}[!htb]
\label{comparison table}
\begin{center}
\begin{tabular}{c|cccc|cccc}
\toprule
\multirow{2}{*}{Method}   & \multicolumn{4}{c}{Translation absolute Error (cm)} & \multicolumn{4}{c}{Rotation absolute Error (${}^\circ $)} \\
   & mean  & X  & Y  & Z   & mean  & Roll  & Pitch & Yaw \\
\midrule
w/o geometric-consistency  & 1.040 & 1.020 & 0.840 & 1.260  & 0.047& 0.034 & 0.042 & 0.066 \\
w/o appearance-consistency &3.796 &4.980 &5.030 &1.380  &0.197 &0.055 &0.035 &0.502  \\
\hline
RobustCalib (\textbf{ours}) &0.806 &0.820  &0.740  &0.860   &0.046&0.034&0.037&0.068 \\ 
\bottomrule
\end{tabular}
\end{center}
\caption{Ablation study results of RobustCalibe on KITTI dataset.}
\end{table*}

\subsubsection{Results on nuScenes.} 
The calibration results for our proposed RobustCalib on nuScenes are shown in Table \ref{tab:compare-nus}. We can observe that RobustCalib achieves a mean translation error of 1.600cm (x, y, z: 2.000cm, 1.530cm, 1.270cm), a mean angle error of $0.114^\circ$ (roll, pitch, yaw: $0.103^\circ$, $0.085^\circ$, $0.155^\circ$). Compared to LCCNet~\cite{lv2021lccnet}, the mean translation and angle error of RobustCalib are reduced by 0.618cm and $0.04^\circ$.

\subsubsection{Results on MT-ADV.}
\label{sec:mtadv-ex}
To further demonstrate the performance of the proposed method, we conducted experiments on the MT-ADV dataset. Compared with KITTI dataset, this dataset is more challenging. In KITTI dataset, our RobustCalib and other high-performance methods only calibrated the LiDAR and the left color camera (i.e., one-to-one). However, in the MT-ADV dataset, in actual operational scenarios, we calibrate one LiDAR with multiple cameras, which better meets the requirements of the practical application scenarios. The proposed method shows general abilities in calibrating different cameras. We achieve a mean translation error of 1.38cm (x, y, z: 1.55cm, 1.45cm, 1.16cm), a mean angle error of $0.103^\circ$ (roll, pitch, yaw: $0.126^\circ$, $0.107^\circ$, $0.077^\circ$).

Except KITTI dataset, we also visualize the qualitative calibration results on MT-ADV dataset, which requires to align one LiDAR to multiple cameras. As shown in Fig. \ref{fig:MT_res}, the proposed RobustCalib shows robust and accurate re-calibration ability under different viewpoints. Specifically, for objects such as trees and poles, the uncalibrated projection results present large offsets. However, our method achieves effective alignment between point clouds and images from different viewpoints.

\subsection{Ablation Study}
\label{sec:ablation}
In this section, the influence of appearance-consistency learning and geometric-consistency learning on the calibration results is explored. We independently disable the appearance-consistency loss and the geometric-consistency loss in the experimental setting. These experiments are conducted on KITTI and the corresponding results are listed in Table. An observation is that when disabling appearance-consistency learning in the calibration process, there appears an obvious performance drop. When the geometric-consistency learning is disabled, there is a slight performance loss. The main reason is that the intensity from LiDAR sensor provides more helpful visual features and patterns, which is more informative than that of depth. When enabling the appearance-consistency learning and geometric-consistency learning together, our method achieves better calibration results.

\section{Conclusion}
In this paper, we present a novel approach to estimate extrinsic calibration between LiDAR and cameras. During the training process, we design an appearance-consistency loss and a geometric-consistency loss for implicitly optimizing the extrinsic parameters. During the inference phase, our pipeline exhibits a more concise architecture within a subnetwork. Given a pair of LiDAR point cloud and camera images, as well as the uncalibrated extrinsic between them, our method directly infers the 6 DOF of the extrinsics calibration. On KITTI dataset, our method yields a mean calibration error of 0.806cm for translation and $0.046^\circ $ for rotation, which achieves a new state-of-the-art. Furthermore, the proposed RobustCaliob still demonstrates superior performance on the MT-ADV dataset, which involves the calibration between LiDAR and multiple surround-view cameras with a larger number of samples. We believe our study will provide valuable insights and contribute to extrinsic calibration for the community. 
\section{Acknowledgement}

This work is supported by National Key R\&D Program of China (No. 2022ZD0118700).

{
    \small
    \bibliographystyle{ieeenat_fullname}
    \bibliography{main}
}

\end{document}